\documentclass[runningheads,a4paper]{llncs}

\usepackage{amssymb}
\usepackage{graphicx}
\usepackage{subfigure}
\usepackage{url}
\usepackage{algorithm}
\usepackage{algorithmic}
\usepackage{booktabs}

\newcommand{\keywords}[1]{\par\addvspace\baselineskip
\noindent\keywordname\enspace\ignorespaces#1}

\begin{document}

\mainmatter  

\title{An STDP-Based Supervised Learning Algorithm for Spiking Neural Networks}
\titlerunning{An STDP-Based Supervised Learning Algorithm}

%
%
\author{Zhanhao Hu\inst{1}\and Tao Wang\inst{2}\and Xiaolin Hu\inst{3}}

\authorrunning{Zhanhao Hu, Tao Wang, Xiaolin Hu}

\institute{Department of Physics, Tsinghua University, Beijing, China\\
\and
Huawei Technology, Beijing, China
\and
Tsinghua National Laboratory for Information Science and Technology (TNList),\\
Department of Computer Science and Technology, \\
Center for Brain-Inspired Computing Research (CBICR),\\
Tsinghua University, Beijing, China\\
}

%
%

\maketitle

\begin{abstract}
	Compared with rate-based artificial neural networks, Spiking Neural Networks (SNN) provide a more biological plausible model for the brain. But how they perform supervised learning remains elusive. Inspired by recent works of Bengio et al., we propose a supervised learning algorithm based on Spike-Timing Dependent Plasticity (STDP) for a hierarchical SNN consisting of Leaky Integrate-and-fire (LIF) neurons. A time window is designed for the presynaptic neuron and only the spikes in this window take part in the STDP updating process. The model is trained on the MNIST dataset. The classification accuracy approach that of a Multilayer Perceptron (MLP) with similar architecture trained by the standard back-propagation algorithm.
\end{abstract}

\section{Introduction}
	Rate-based deep neural networks (RDNN) with back-propagation (BP) algorithm have got great developments in recent years \cite{lecun2015deep}. Neurons in these networks deliver information by floating numbers. But in the brain, signals are carried on by spikes, a kind of binary signals. This property can be captured by a spiking neural networks (SNN). But how the SNNs are trained remains largely unknown.
	
	Several works studying supervised algorithm on SNN have made some progress recently. Some works \cite{bohte2002error}\cite{booij2005gradient}\cite{ghosh2009new}\cite{gutig2006tempotron}\cite{ponulak2010supervised}\cite{xie2017efficient} make use of time coding by spikes. In a very first work \cite{bohte2002error}, each neuron is only allowed to fire a single spike. The model is then expanded to allowing multiple spikes by later studies \cite{booij2005gradient}\cite{ghosh2009new}. Networks in these papers usually need to keep multiple channels with independent weights between two neurons. These channels account for different time delays \cite{bohte2002error} or order numbers of spikes in the spike train \cite{xie2017efficient}. These algorithms are designed to learn spike trains, but classification on large datasets is hard for these models. In fact, the algorithms need to convey real numbers to spike trains. Due to the difficulty of this conversion and recognizing ability of the network, these models can only work on very simple datasets.

	Recently, Bengio et al. proposes an idea to build a two-phased learning algorithm for energy-based models called e-prop \cite{Scellier2017Equilibrium}. They implement the algorithm on an energy-based model with input neurons clamped to input data and output neuron variable under target signals. Neurons are free from target signals in the first phase, and the state of which is denoted by $s^0$. Dynamics of output neurons are changed slightly by target signals in the second phase, and the state of neurons is $s^\xi$. Let $\rho()$ represent the active function. The weight $W_{ij}$ of synapse between neuron $j$ and neuron $i$ is updated by
	\begin{equation}
        W_{ij}\gets W_{ij}+\eta\Delta W_{ij} \; ,
    \end{equation}
    where
	\begin{equation}
        \Delta W_{ij}\propto \lim_{\xi \to 0} \frac{1}{\xi}(\rho(s_i^\xi)\rho(s_j^\xi)-\rho(s_i^0)\rho(s_j^0)) \; .
        \label{equ:learn_rate_sym}
    \end{equation}
    And this rule is a symmetric version of another rule
    \begin{equation}
        \Delta W \propto \dot{s_i}\rho(s_j) \; ,
        \label{equ:learn_rate_asy}
    \end{equation}
    which is studied in previous work \cite{Bengio2015STDP}. In the work a link has been made between (\ref{equ:learn_rate_asy}) and Spike-Timing Dependent Plasticity (STDP) rule.

    STDP rule is thought to be an ideal basis of algorithms on SNN. It is first found in physiological experiment \cite{Markram1997Regulation}, defines that the plasticity of a synapse is only dependent on the time difference of spikes from the two neurons attached by this synapse. But computational significance of STDP is not clear. Several works implement STDP on learning algorithms \cite{querlioz2013immunity}\cite{diehl2015unsupervised}\cite{kheradpisheh2016stdp}. They all take an idea that utilizing the simple property that the synaptic weight is strengthened when the postsynaptic spike is after the presynaptic spike, thus a strict order of presynaptic and postsynaptic spike is needed.

    In this work we propose a new STDP-based algorithm on SNN. Also, we find that simply computing all of the spikes using the STDP rule results in poor results. We modify the spike pairs that perform STDP rule, and achieve good results on same benchmark image classification dataset. We stress that we do not change the original STDP rule on single pair of spikes, but provide a way that how to use the STDP rule.

\section{Method}
	\subsection{The network}
		The network is a bidirectionally connected network with asymmetric weights based on the leaky integrate-and-fire (LIF) neuron model \cite{dayan2001theoretical}. The state of neuron $i$ is described by membrane potential $V_i$. The dynamics of $V_i$ is:
		\begin{equation}
			\tau_V\frac{\mathrm{d}V_i}{\mathrm{d}t}=-V_i+E_L-r_m\sum_{j\in \Gamma_i}\bar{g}_{s,ij} P_{s,j}(V_i-E_{s,j})+R_mI_e\
            \label{equ:ori_V} \; , 	
        \end{equation}
		where $I_e$ is input current, $E_L$ is equilibrium potential, $E_{s,j}$ is determined by types of neurotransmitter, $\tau_V$ is a time constant, and $R_m$ is a resistance constant. $\Gamma_i$ is the set of neurons that have synapses to neuron $i$. The membrane potential $V_i$ triggers the neuron to release a spike when it reaches a threshold $V_{th}$, and then is reseted to $V_{reset}$ after the spike. $\bar{g}_{s,ij}P_{s,j}$ represents synaptic conductance from neuron $j$ to neuron $i$, where $\bar{g}_{s,ij}$ represents the maximum strength of the synapse, and $P_{s,j}$ represents the probability of opened neurotransmitter gates. The dynamics of $P_{s,j}$ is
		\begin{equation}
			\tau_P\frac{\mathrm{d}P_{s,j}}{\mathrm{d}t}=-P_{s,j}+\sum_k\delta(t-T_j^{(k)}) \; .
            \label{equ:ori_P}
        \end{equation}
        The variable $P_{s,j}$ increases by a unit amount every time neuron $j$ spikes, and decreases to zero spontaneously. $\delta()$ is a Dirac function, which means $\delta(x)=0,(x\neq0)$ and $\int_{-\infty}^{\infty} \delta(x)\mathrm{x}=1$. $[T_j^{(1)},T_j^{(2)},...]$ represents for spike train of neuron $j$.

        We set all $E_{s,j}$ to $0\;V$ , and the input current $I_e$ to $0\;\mu A$. Also, for the sake of convenience, we write $\bar{g}_{s,ij}$ to $W_{ij}$, and introduce an input summation for postsynaptic neuron $i$
        \begin{equation}
            P_i=\sum_{j\in \Gamma_i}\bar{g}_{s,ij} P_{s,j}\; ,
        \end{equation}
        and rewrite the basic dynamics (\ref{equ:ori_V}) and (\ref{equ:ori_P}) as
        \begin{equation}
            \tau_V\frac{\mathrm{d}V_i}{\mathrm{d}t}=-V_i+E_L-r_mP_i(V_i-E_s)\; ,
            \label{equ:our_V}
        \end{equation}
		\begin{equation}
            \tau_P\frac{\mathrm{d}P_i}{\mathrm{d}t}=-P_i+\sum_{j,k}W_{ij}\delta(t-T_j^{(k)})\; .
            \label{equ:our_P}
        \end{equation}

        The network consists of an input layer, a hidden layer, and an output layer. We denote the data for supervised learning by normalized input signal $v_x$ and target signal $v_y$.

        For neuron $i$ in the input layer, we simply let it be controlled by input signal $v_{x,i}$:
		\begin{equation}
            P_i=P_0v_{x,i}\; ,
            \label{equ:con_x}
        \end{equation}
        and $P_0$ is a constant to convert the scales. Neurons in the input layer fire in a fixed pattern under an input proportional to input signal. The neurons in the hidden layer are not affected by any signals from data directly, and act according to (\ref{equ:our_V}) and (\ref{equ:our_P}).

        Situation for neurons in the output layer is a bit more complicated. Like the e-prop method \cite{Scellier2017Equilibrium}, learning is performed in two phases, named inference phase and learning phase in this paper. The only difference between the two phases is the dynamics of the output layer neurons. In the inference phase, the neurons also act according to (\ref{equ:our_V}) and (\ref{equ:our_P}). The network gives an inference result by counting the frequency of spikes of output neurons in this phase. And in the learning phase, we add an item represents for effect of target signals:
        \begin{equation}
            \tau_P\frac{\mathrm{d}P_i}{\mathrm{d}t}=-P_i+\sum_{j,k}W_{ij}\delta(t-T_j^{(k)})+\beta(P_0v_{y,i}-P_i)\; ,
            \label{equ:con_y}
        \end{equation}
        where $v_{y,i}$ is the $i$th target signal and $\beta$ controls the effectiveness of target signals.

    \subsection{The learning rule}
\begin{figure}[htbp]
	    	\centering
	    	\subfigure[]{
	        \includegraphics[width=0.4\textwidth]{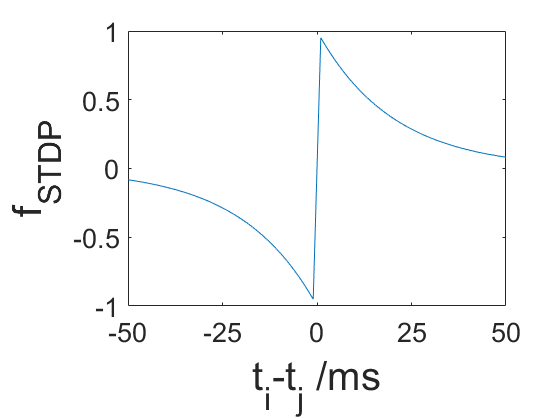}
	        }
	        \subfigure[]{
	        \includegraphics[width=0.4\textwidth]{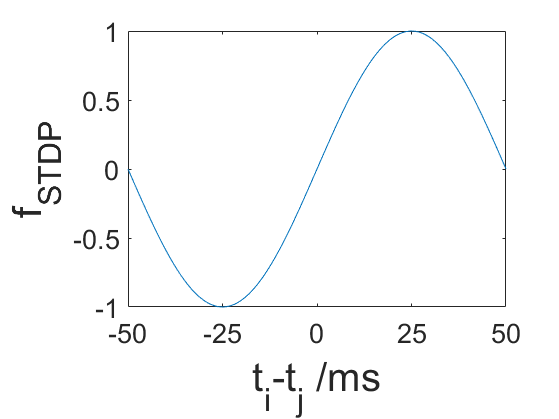}
	        }
	        \subfigure[]{
	        \includegraphics[width=0.6\textwidth]{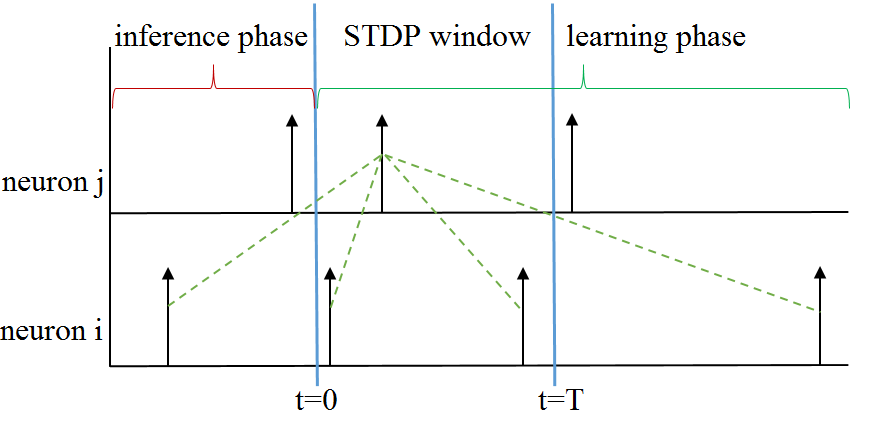}
	        }
	        \caption{(a) The exponential form of STDP function. (b) The sinusoidal form of STDP function. (c) Illustration of details of implementing STDP. The learning window is embedded in the learning phase, which is the space between the two vertical lines. Only spike pairs indicated by dotted lines are taken into consider for updating $W_{ij}$, which is the synaptic weight from presynaptic neuron $j$ to postsynaptic neuron $i$.}
        	\label{fig:STDP}
    	\end{figure}

    	We adopt the original STDP functions. The STDP function represent the relationship of modification $\delta W_{ij}$ of the synapse $W_{ij}$ from presynaptic neuron $j$ to postsynaptic neuron $i$, and the firing time of two spikes fire at $t_j$ and $t_i$ respectively. The commonly used exponential form \cite{song2000competitive} of STDP function can be
        \begin{equation}
            \delta W_{ij}(t_i,t_j)=f(t_i-t_j)=\left\{
            \begin{array}{cl}
                e^{-\frac{t_i-t_j}{\tau_{m}}}&\mbox{, when }t_i>t_j \\
                0&\mbox{, when }t_i=t_j  \\
                -e^{-\frac{t_j-t_i}{\tau_{m}}}&\mbox{, when }t_i<t_j \\
            \end{array}\right.\; .
            \label{equ:STDP_exp}
        \end{equation}
        And we can also use a sinusoidal form \cite{xie2000spike} as
        \begin{equation}
            \delta W_{ij}(t_i,t_j)=f(t_i-t_j)=\left\{
            \begin{array}{cl}
                \sin(\frac{t_i-t_j}{\tau_w}\pi)&\mbox{, when }\Delta t\in [-\tau_w,\tau_w] \\
                0&\mbox{, otherwise} \\
            \end{array}\right.\; ,
            \label{equ:STDP_sin}
        \end{equation}
        where $\tau_w$ is a time constant. The two functions are plotted in Fig.~\ref{fig:STDP}. During the experiment we found that the sinusoidal form resulted in better consequence, so all results presented in the paper are based on (\ref{equ:STDP_sin}).

        The STDP rule is implemented on a time window in the learning phase after the inference phase. We find that simply summing up all of the spike pairs in a bidirectional network does not work. When the STDP function is approximately anti-symmetric which means $f(\delta t)=-f(-\delta t)$, we have
        \begin{equation}
            \Delta W_{ij}=\sum_{t_i,t_j}f(t_i-t_j)=-\sum_{t_i,t_j}f(t_j-t_i)=-\Delta W_{ji} \; .
            \label{equ:asy_delta}
        \end{equation}
        It means that the synapse modifications in two directions of two neurons are always opposite. Consider a situation that two neurons' firing rates are increasing in a same mode, so that average modification of the two synapse are expected to be symmetric, which is $\Delta W_{ij}=\Delta W_{ji}$. Along with (\ref{equ:asy_delta}), we have $\Delta W_{ij}=\Delta W_{ji}=0$. This makes no sense for learning and implementing this operation can not learn the model well.

        For breaking this symmetry we made a slight modification. We redefined the rules of multiple spikes in a time window $[0,T]$. That is, for synapse $W_{ij}$, spikes fired by presynaptic neuron $j$ only in time window $[0,T]$, and spikes fired by postsynaptic neuron $i$ in time window $[-\infty,\infty]$ are taken into account:
        \begin{equation}
        	\Delta W_{ij} \propto \int_0^T\mathrm{d}t_j\int_{-\infty}^\infty\mathrm{d}t_if(t_j-t_i)\sum_{k,l}\delta(t_i-T_i^{(k)})\delta(t_j-T_j^{(j)}) .	
        \end{equation}
        Because of the local property of STDP rule, which means only spikes that the time distance is not larger than $\tau_w$ in (\ref{equ:STDP_sin}) actually effect, the scope of spikes fired by postsynaptic neuron is $[-\tau_w,T+\tau_w]$ in fact.

        In fact, when STDP rule is implemented on time window $[0,T]$, it means the STDP is somehow "turned on" at the time $t=0$ and "turned off" at the time $t=T$. And more specifically, STDP can be considered as a consequence of some kinds of biochemical signals from both presynaptic neuron and postsynaptic neuron \cite{clopath2010connectivity}. We propose an idea that STDP is considered to be "turned on" by activating the production or transmission of the biochemical signal triggered by presynaptic neuron spikes, and also it is "turned" off by suppressing these signals, while signals related to postsynaptic spikes are existed all time along.

        The learning algorithm is summarized in Algorithm \ref{alg:1}.
        \begin{algorithm}[!h]
    		\caption{Training the spiking neural network}
    		\label{alg:1}
    		\begin{algorithmic}[1]
        		\STATE Simulate an inference phase in the time window of $[-t_0,0]$ and record the spike train $[T_i^{(1)},T_i^{(2)},...,T_i^{(m_i)}]$.
        		\STATE Simulate a learning phase in the time window of $[0,T+t_w]$ and record the spike train $[T_i^{(m_i+1)},T_i^{(m_i+2)},...,T_i^{(M_i)}]$.
        		\STATE $W_{ij}^{(n+1)}=W_{ij}^{(n)}+\alpha \int_0^T\mathrm{d}t_j\int_{-\infty}^\infty\mathrm{d}t_if(t_j-t_i)\sum_{k,l}\delta(t_i-T_i^{(k)})\delta(t_j-T_j^{(j)})$.
    		\end{algorithmic}
		\end{algorithm}

\section{Results}
	
	We implement the model on the MNIST dataset. The dataset contains 60,000 training images and 10,000 test images. And the images are in gray scale and have size $28 \times 28$. The size of the network is 784-200-10, which indicates the numbers of neurons in input layer, hidden layer, and output layer, respectively.

    \begin{figure}[htbp]
    	\centering
    	\subfigure[Input summations]{
        \includegraphics[width=0.4\textwidth]{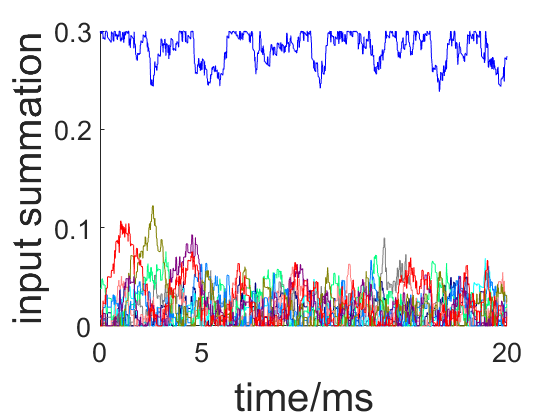}}
        \subfigure[Membrane potential]{
        \includegraphics[width=0.4\textwidth]{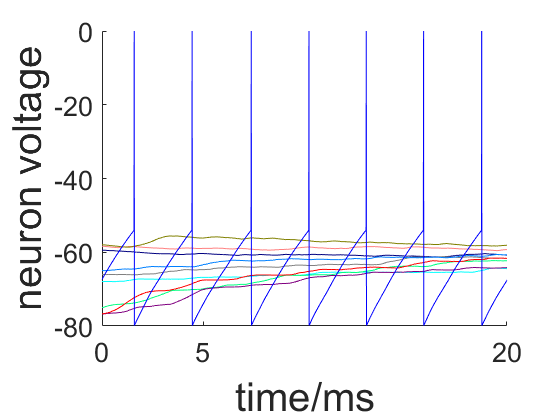}
        }
        \subfigure[Input summation]{
        \includegraphics[width=0.4\textwidth]{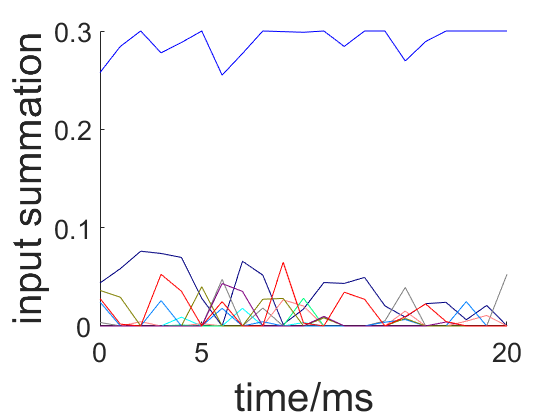}
        }
        \subfigure[Membrane potential]{
        \includegraphics[width=0.4\textwidth]{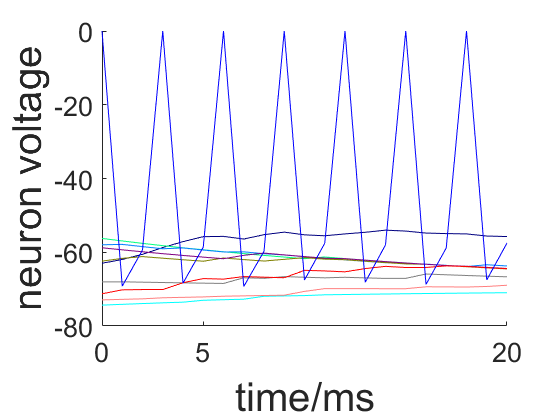}
        }
        \caption{Simulation illustration of input summation and membrane potential of 10 output layer neurons in an inference phase. The synaptic weights have been trained on MNIST dataset. Only the 8th neuron have a maximal input, and fires in a maximal pattern, while other neurons are not. The simulation for (\ref{equ:our_V}) and (\ref{equ:our_P}) is processed using Euler method. We have tried different time steps, as 0.01 ms for (a)(b) and 1 ms for (c)(d)}
        \label{fig:spike_0.01}
    \end{figure}

    We use the Euler method to approximate the differential function (\ref{equ:our_V}) and (\ref{equ:our_P}). Fig.~\ref{fig:spike_0.01} is the simulation illustration of input summation and the membrane potential of 10 output layer neurons with different time step. We set $\tau_V=20ms,\; \tau_P=10ms,\; E_s=0,\; E_L=-70mV,\; V_{reset}=-80mV,\; V_{th}=-54mV$, and a hard bound $[0, 0.3]$ for $P_i$. In fact, we find that a simulation time step of 1 ms is enough to depict the spiking trains, so we use a step of 1 ms in our later experiment.

    \begin{figure}[htbp]
    	\centering
        \subfigure[]{
        \includegraphics[width=0.4\textwidth]{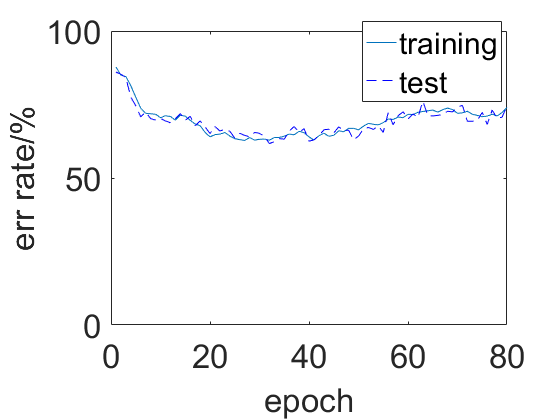}
        }
        \subfigure[]{
        \includegraphics[width=0.4\textwidth]{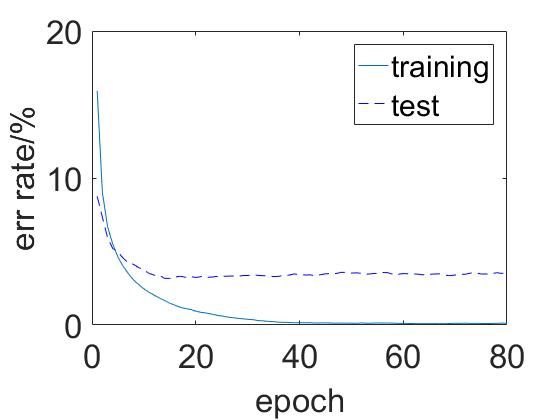}
        }
        \caption{The error rates against epochs on the MNIST dataset. (a) Simply take all of the spike pairs into account. (b) Use the proposed method.}
        \label{fig:err}
    \end{figure}

    We test our model on the MNIST dataset (Fig.~\ref{fig:err}). Using the STDP rule with all of the spikes in the time window taken in to account did not work. By using the proposed method, the error rate on training set is able to decrease to 0.0\% in the experiment, which proves the convergence of algorithm experimentally. For comparison, we also implement the e-prop and MLP which have similar architecture to our model (the same number of input, hidden and output neurons). Several other STDP-based algorithms are also compared. The test accuracies on the MNIST dataset are summarized in Table \ref{tab:test}.  The test accuracy of our method is greater than other STDP-based algorithms, except for the algorithm that use a convolutional architecture \cite{kheradpisheh2016stdp}.

    \begin{table}[ht]
        \centering
        \caption{Comparison of different algorithms}
        \label{tab:test}
        \begin{tabular}{lcc}
            \toprule
            \textbf{Model} & \textbf{Neural coding } & \textbf{ Test accuracy/$\%$}\\
            \midrule
            \textbf{E-prop} & Rate-based & 97.5\\
            \textbf{MLP} & Rate-based & 98.5\\
            \midrule
            \textbf{Two layer network \cite{querlioz2013immunity}} & Spike-based & 93.5\\
            \textbf{Two layer network \cite{diehl2015unsupervised}} & Spike-based & 95.0\\
            \textbf{Convolutional SDNN \cite{kheradpisheh2016stdp}} & Spike-based & 98.4\\
            \textbf{Proposed model} & Spike-based & 96.8\\
            \bottomrule
        \end{tabular}
    \end{table}

\section{Discussion}

	We describe an STDP-based supervised learning algorithm on SNN, and get good results on the MNIST classification task. The accuracy approaches that of an MLP with a similar architecture, which indicates the effectiveness of this algorithm. Compared with existing algorithms for training SNNs, the proposed algorithm have achieved competing results.

	The algorithm suggests that biological neurons may not modify their synapses under the STDP rule all the time. STDP takes effect only when the supervisory signals are applied. In addition, the algorithm suggests that not all spikes of the presynaptic neuron participate in the STDP learning process for the synapse. Instead, there may exist a time window and only the spikes during this window should be counted. But biochemical evidence is needed to validate these predictions.

\section*{Acknowledgment}

	This work was supported in part by the National Natural Science Foundation of China under Grant 91420201, Grant 61332007, Grant 61621136008 and Grant 61620106010, in part by the Beijing Municipal Science and Technology Commission under Grant Z161100000216126, and in part by Huawei Technology under Contract YB2015120018.

\bibliographystyle{splncs03}
\bibliography{main}

\end{document}